# Leveraging Chemistry Foundation Models to Facilitate Structure Focused Retrieval Augmented Generation in Multi-Agent Workflows for Catalyst and Materials Design

Nathaniel H. Park,[1,*] Tiffany J. Callahan,[1] James L. Hedrick,[1] Tim Erdmann,[1] and Sara Capponi[1]

[1]IBM Research–Almaden, 650 Harry Rd. San Jose, CA 95120

*Corresponding author. Email: npark@us.ibm.com

## Abstract

Molecular property prediction and generative design via deep learning models has been the subject of intense research given its potential to accelerate development of new, high-performance materials. More recently, these workflows have been significantly augmented with the advent of large language models (LLMs) and systems of autonomous agents capable of utilizing pre-trained models to make predictions in the context of more complex research tasks. While effective, there is still room for substantial improvement within agentic systems on the retrieval of salient information for material design tasks. Within this context, alternative uses of predictive deep learning models, such as leveraging their latent representations to facilitate cross-modal retrieval augmented generation within agentic systems for task-specific materials design, has remained unexplored. Herein, we demonstrate that large, pre-trained chemistry foundation models can serve as a basis for enabling structure-focused, semantic chemistry information retrieval for both small-molecules, complex polymeric materials, and reactions. Additionally, we show the use of chemistry foundation models in conjunction with multi-modal models such as OpenCLIP facilitate unprecedented queries and information retrieval across multiple characterization data domains. Finally, we demonstrate the integration of these models within multi-agent systems to facilitate structure and topological-based natural language queries and information retrieval for different research tasks.

## Introduction

The advent of large language models (LLMs) with capabilities for tool use has enabled a dramatic expansion of LLM-based systems across chemistry and materials domains. These include applications of chatbots or multi-agent workflows to metal-organic frameworks,[1,2] inorganic materials,[3] automated synthesis,[4] organic synthesis,[5] and protein discovery.[6] Besides the underlying agents, some of the most critical components of these workflows are the tools made available to individual agents. This can include access to retrosynthetic software,[5] predictive models or computational tools,[3] or generating task-specific code for automated experimentation[4]—all of which provide salient information needed by an agent to complete a given task. These examples are frequently used in combination with standard retrieval augmented generation (RAG) methodologies, where an embedded natural-language query is executed against a vector database of embedded documents to retrieve the most relevant documents and provide them to the LLM as part of its prompt.[7–9] As with other types of agent tools, this approach is effective in enhancing LLM performance for answering a particular question by providing the LLM with context it was not trained on. However, it requires significant optimization with regards to the choice of embedding model, chunk size, LLM and its available context window, question rephrasing, use of metadata filtering, and similarity metrics used for retrieval.[8,9]

In many of the agent-based workflows noted above, the implemented tools may utilize structural information in the form of a SMILES string or molecular formula when performing a computation or property prediction,[6,10,11] but not directly in RAG operations focused on structural similarity. Retrieval of relevant information based on structural similarity, whether focused small-molecules, materials, or reactions, is one of the most critical tasks during any chemistry or materials research endeavor. Therefore, enabling researchers to use natural-language accompanied by chemical language to query structure-linked information resources would provide a powerful augmentation of LLM capabilities. Structure-based RAG operations necessitate a vector-based representation of a given compound or material whereby similarity queries may be conducted. While there are a significant variety of molecular fingerprints available through cheminformatics packages such as RDKit,[12] the use chemistry language foundation models is highly attractive due to their potential dual use within an agentic system as an embedding model and a predictive tool. Here, the same model would be able to support both prediction tasks, such as providing accurate assessments of properties for a compound of interest and support semantic structural queries for relevant information based on a compounds structure or structural similarity. Despite these advantages, there exist only a few reports on the utilization of deep learning chemistry models as embedding models for similarity-based searches—limited largely to organic molecules and inorganic materials.[13,14] And while the LLM systems for chemistry applications have been demonstrated to search APIs such as PubChem or the general internet,[5] these search interfaces do not offer the potential breadth of both query options or structural similarity features that may be possible when using a chemistry foundation model to embed structural information and with relevant metadata. Additionally, queries based on other modalities, such as images of characterization data (e.g., NMR spectra or GPC traces), have largely not been evaluated within an LLM-based agentic system nor within traditional chemistry database interfaces in spite of the immense benefit such features would confer to researchers.

To overcome these limitations and dramatically expand the capabilities of multi-agent systems for materials design and development tasks, we surmised that: 1) a single, high-performance chemistry foundation model could be adapted to facilitate semantic structure searches across small-molecules, polymers, and reactions; 2) such a model could be used in combination with a multimodal embedding model to enable semantic queries over images and chemical structures; and 3) these systems can be integrated within multi-agent systems to provide richer context while performing materials design tasks. Before building a complete semantic structure search-based multi-agent RAG workflow, it was imperative to evaluate the effectiveness of the latent representations generated by the chemistry language foundation models for structural similarity queries with many competent models that may fill such a role. Prior work focused on the use of ChemBERTa models to enable similarity searches for functional analogues of drugs,[14–16] however there are other potential models that could work including Mol-BERT,[17] Mol2Vec,[18] GPT-MolBERTa,[19] or ChemGPT.[20] For our investigation, we selected MoLFormer as a baseline embedding model which has recently been shown to be highly performant across numerous benchmarks from MoleculeNet,[21] effective at capturing molecular similarity, as well as potentially learning 3D spatial relationships from SMILES inputs.[22] To evaluate structural similarity queries, we compiled a focused dataset of ~2.5M organic small-molecules using open-source data and historical data from our own work.[23–26] The SMILES representations were canonicalized and vector embeddings were computed for each compound using MoLFormer before insertion into a Milvus vector database (see Supporting Information for more details).[27] Using this dataset we evaluated the ability of MoLFomer embeddings to facilitate retrieval of structurally similar compounds from sample queries on known organocatalysts for ring-opening polymerization (ROP) using the Milvus vector search capabilities (Fig. 1).[28,29]

For the result of each query, several metrics relating to the structural similarity based on either MoLFormer embedding distance (cosine, Euclidean, Fig.1) or fingerprints for each compound generated using RDKit (Tanimoto, RDKit, MACC, and Dice, Fig 1). Each score represents the similarity between the resulting compound and the query compound. For 1,8-diazabicyclo[5.4.0]undec-7-ene (DBU) **1a**, the top result was itself followed by several closely related structural analogues based on the cosine or Euclidean similarity (Fig. 1A). Interestingly, the second rank compound **1b** was given a much lower similarity scores from most fingerprint derived metrics despite differing from **1a** only in the saturation of the imine (Fig. 1A). For **2a**, a organocatalyst for ROP designed with the aid of generative AI,[29] we find that while the original query was not present in the dataset, most of the core functional group features—an cyclic 5-membered guanidine and an aromatic ring—based on visual inspection (Fig. 2B). As with **1a**, most of the fingerprint-based similarity metrics indicate the result compounds as highly dissimilar, with exception of MACC (Fig. 2B). Overall, these examples further substantiate prior results[22] demonstrating MoLFormer embeddings do indeed capture relevant structural information and similarity despite differing substantially in several cases from fingerprint-based metrics.

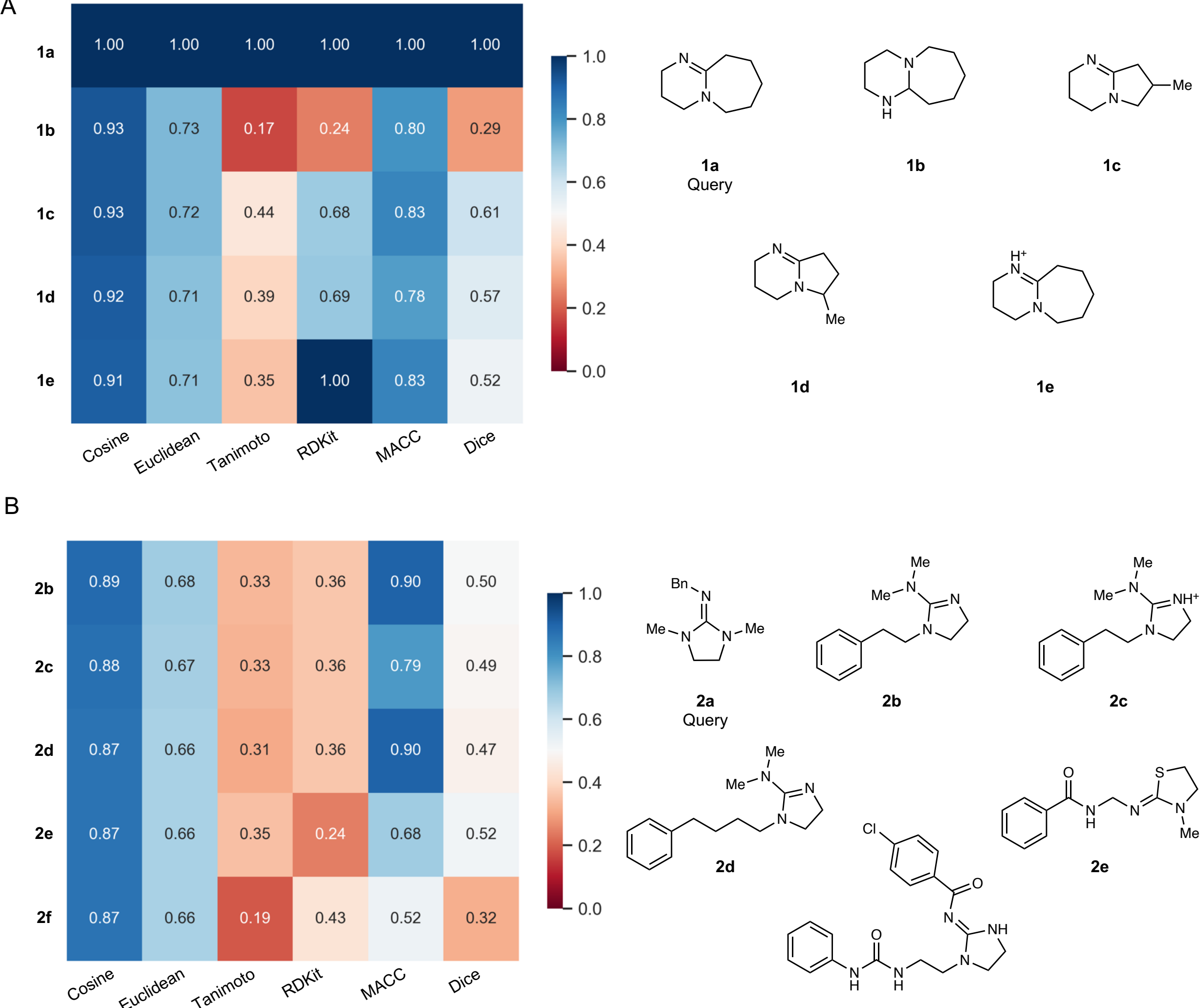

**Figure 1.** A) Heatmap of similarity metrics for the results of the query of **1a** against a small-molecule collection. Structures of each of the top five closest compounds are shown beneath the heatmap. B) Heatmap of similarity queries of **2a** against a small-molecule collection. Structures of each of the top five closest compounds. For the similarity metrics displayed in the heatmap, cosine and Euclidean refer to the respective distance between the query embedding and the result embedding converted into similarity scores. Tanimoto, RDKit, MACC, and Dice refer to similarity scores from molecular fingerprints generated using the RDKit package for the query and returned compound. For all metrics, a value closer to 1 indicates high similarity whereas a value closer to zero indicates lower similarity. See Supplementary Information for details.

With the success of the similarity queries using MoLFormer embeddings in conjunction with a Milvus vector store, we sought next to expand search capabilities beyond simple small-molecule queries to maximize the diversity of search options available to LLM agents. In natural language, differences in word vector embeddings have been demonstrated to capture lexical relationships between different words, with classic examples of *King – Man + Woman = Queen* or *Paris – France + Poland = Warsaw* capturing the relationships of gender and capital city.[30] Given that MoLFormer is a chemistry language model based on SMILES syntax, we surmised that differences between vector embeddings between two compounds within the base model latent space should correspond to differences in structure (Fig. 1B). Moreover, this implies that combinations of vector embeddings from compounds of interest can be used to identify novel analogues based on functional group features of both compounds as well as utilize scalar property values to manipulate the magnitude of the vector—facilitating unique similarity queries to access different sets of data. To implement these ideas, we first created an identical collection of ~2.5M molecules used in Fig. 1 where instead each vector was now scaled by the compounds' corresponding molecular weight. This was tested on guanidine **3a**, which was used as a search input to either the molecular weight scaled collection, or the base collection used previously (Fig. 2A). The results in Fig. 2A demonstrated the success of both approaches in retrieving structurally relevant compounds, however, the metadata filtering is limited to identical top results within certain molecular weight ranges (**3f**, Fig. 2A)—potentially requiring different filtering strategies depending on the desired output.

In addition to utilizing the vector magnitude to influence similarity search results, we also investigated whether the addition and subtraction of vectors corresponding to different functional groups resulted in similar lexical relationships as observed with examples like: *King – Man + Woman = Queen*. This approach was evaluated on two catalysts **4a** and **4b** where their corresponding vector embeddings were subjected such operations. With **4a** the subtraction of the vector corresponding to dimethylurea and the addition of dimethylthiourea resulted in a vector embedding that provided corresponding thioureas (**4c**) when queried against the database (Fig. 2B). With **4b** the results of similar operation were less obviously successful, but this is anticipated behavior based on a limited collection of molecules (Fig. 2B). Nonetheless, the results did provide similar examples with fluorine containing amines. Finally, the vector embeddings of **4a** and **4b** were averaged and queried against the database with the top results bearing the structural features of both the parent molecules (Fig. 2C).

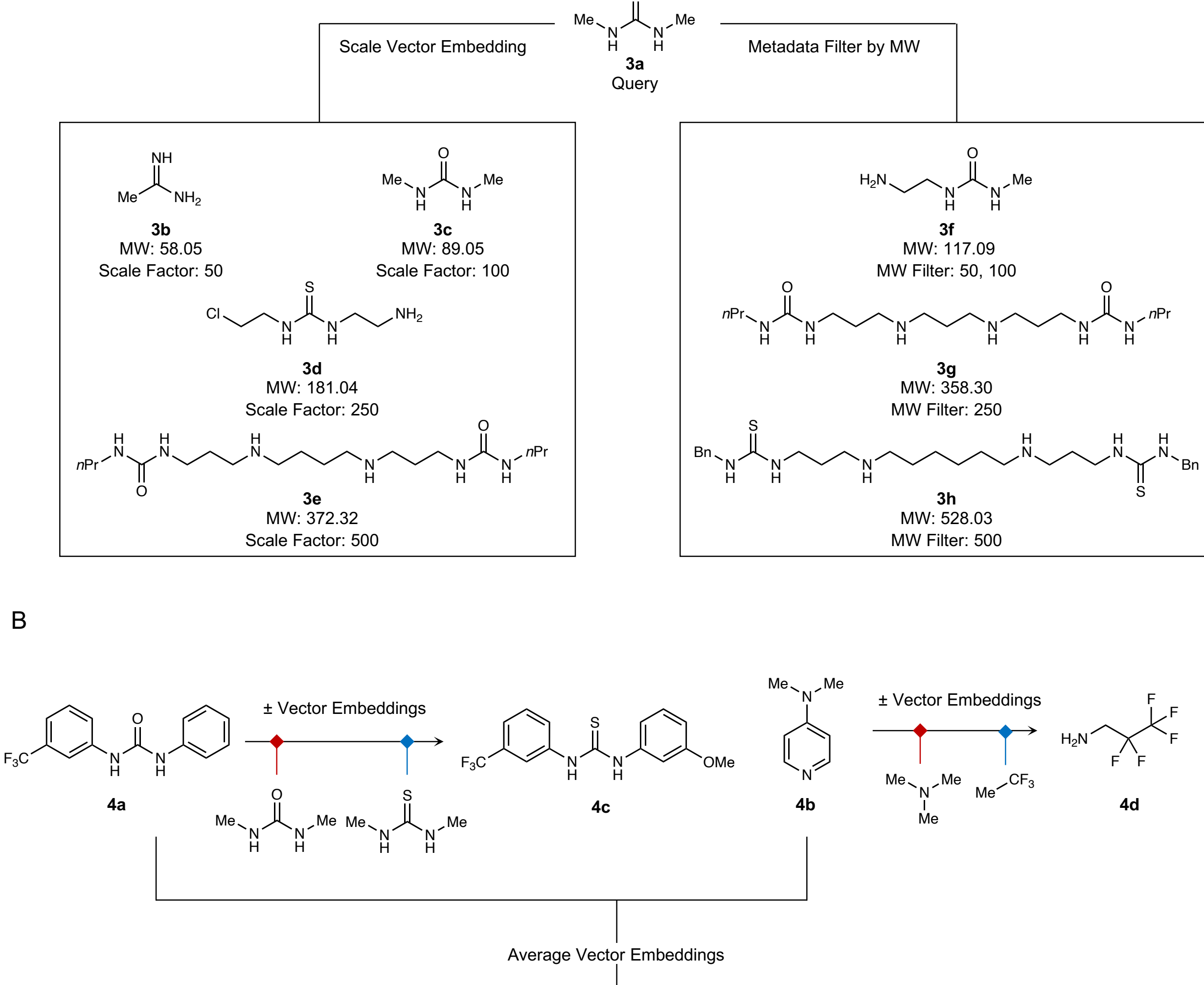

**Figure 2.** A) Comparison of results obtained for queries using run using metadata filters compared to normalized vector embeddings scaled by a target molecular weight. B) Schematic depiction of search results obtained through mathematical manipulation of query vector embeddings through addition-subtraction of corresponding functional group vectors or averaging two queries.

Having demonstrated the effectiveness of using MoLFormer vector embeddings on a variety of different similarity search strategies, we next sought to evaluate their feasibility towards polymers and reactions. MoLFormer was trained compounds with SMILES strings containing less than 200 tokens were filtered from the pre-training dataset, precluding large SMILES strings which may potentially represent macromolecules.[22] Outside of biological materials, such as peptides, nucleic acids or large natural products, macromolecules and other complex materials often exhibit stochastic features relating to their structure and are hence poorly represented by discrete, one-dimensional SMILES strings. This problem is frequently circumvented in deep learning models through a reduction in the complexity by treating polymers as single, discrete repeating units via SMILES strings with variable attachment points denoted by the asterisk character.[31,32] While this approach tends to overlook the stochastic nature of materials as well as neglects to account for end groups or more complex polymer topologies, it does produce systems capable of providing predicted values for polymer properties within these restrictions.[31,32] As the tokenizer for MoLFormer covers the entire SMILES grammar,[22,33] including the use of special tokens such as the use of asterisks for variable attachment points, we hypothesized that it could also serve as a suitable source of embeddings to facilitate queries based on both polymer structural and topological similarity. It was unclear, however, as to whether these latent embeddings would capture the same level of molecular similarity given SMILES fragments with asterisks were unlikely to have been part of any pre-training set for the model. With this in mind, we leveraged dot-separated SMILES strings containing asterisks to represent repeat units or other polymer components to evaluate their suitability for use in semantic queries for polymeric materials.

To evaluate the use of MoLFormer embeddings applied to polymers, we first curated a set of polymer data from both open literature and our own historical data totaling in ~2.5M SMILES strings representing predominantly homopolymers.[34–39] As with the small-molecule embeddings, these were inserted into a Milvus collection, against which similarity queries may be run. First, we tested the embeddings on very simple queries such as polyimide **S4a** and polystyrene **S4c**, both of which returned sets of highly similar compounds (Fig. S3A). Averaging the embeddings of **S4a** and **S4c** enabled the search to identify features of both, consistent with results observed on the small-molecules and again indicative of the ability of MoLFormer embeddings to capture semantic structural relationships.

Next, we investigated the possibility of utilizing additional scalar weights with the component vector embeddings to enable not only querying for structural similarity but also topological and molecular weight similarity. In this approach, the polymer structure is broken into its structural components, each of which is embedded independently using MoLFormer (Fig. 3A). The individual embeddings are then subjected to weighted averaging method to afford the final vector embedding for a polymer (Fig. 3A). To evaluate this approach, we generated a synthetic dataset consisting of both ROP and acrylate polymers (~2M) with randomly generated molecular weights, dispersity, and degrees of polymerization. Two approaches for computing the final polymer embedding from the component embeddings were investigated. The first approach averaged the individual component vectors and then scaled the averaged embedding by the dispersity and the $\log_{10}$ of the $M_n$ (see Supplementary Information). The second approach focused on the log weighted average of each of the structural components' contribution to the overall $M_n$ of the polymer. Two separate collections from synthetic polymer data were constructed utilizing these embedding approaches and the same query polymer (**5a**, Fig. 3A) was executed against both collections using their respective embedding methods. For the first embedding approach (Fig. 3B), the returned results exhibited high cosine similarity yet diverged strongly in terms of Euclidean similarity measures (Fig. 3B). This can readily be seen within returned results with **5b** being identical structurally but differing more starkly in terms of molecular weight (Fig. 3B). The next closest compounds **5c-1** and **5c-2** exhibited some similarity in structure and in terms of molecular weight. In contrast, the second embedding approach provided results with high cosine and Euclidean similarity (Fig. 3C). The first three results (**5f**) were both structurally identical to the

query—with the top result (**5f-1**) being closest in $M_n$—and the only examples of **5a** in the entire collection (Fig. 3C). Additionally, following results both contained the same structural components within the polymer (**5g** and **5h**, Fig. 3C), as opposed to the other embedding methodology, where only the methanol initiator component was conserved across all the top results (Fig. 3D). Overall, these results demonstrate both the strong influence of the methodology used to scale embeddings on the returned results as well as the ability to use embeddings in combination with a scalar of property data to facilitate unprecedented similarity queries for polymeric materials.

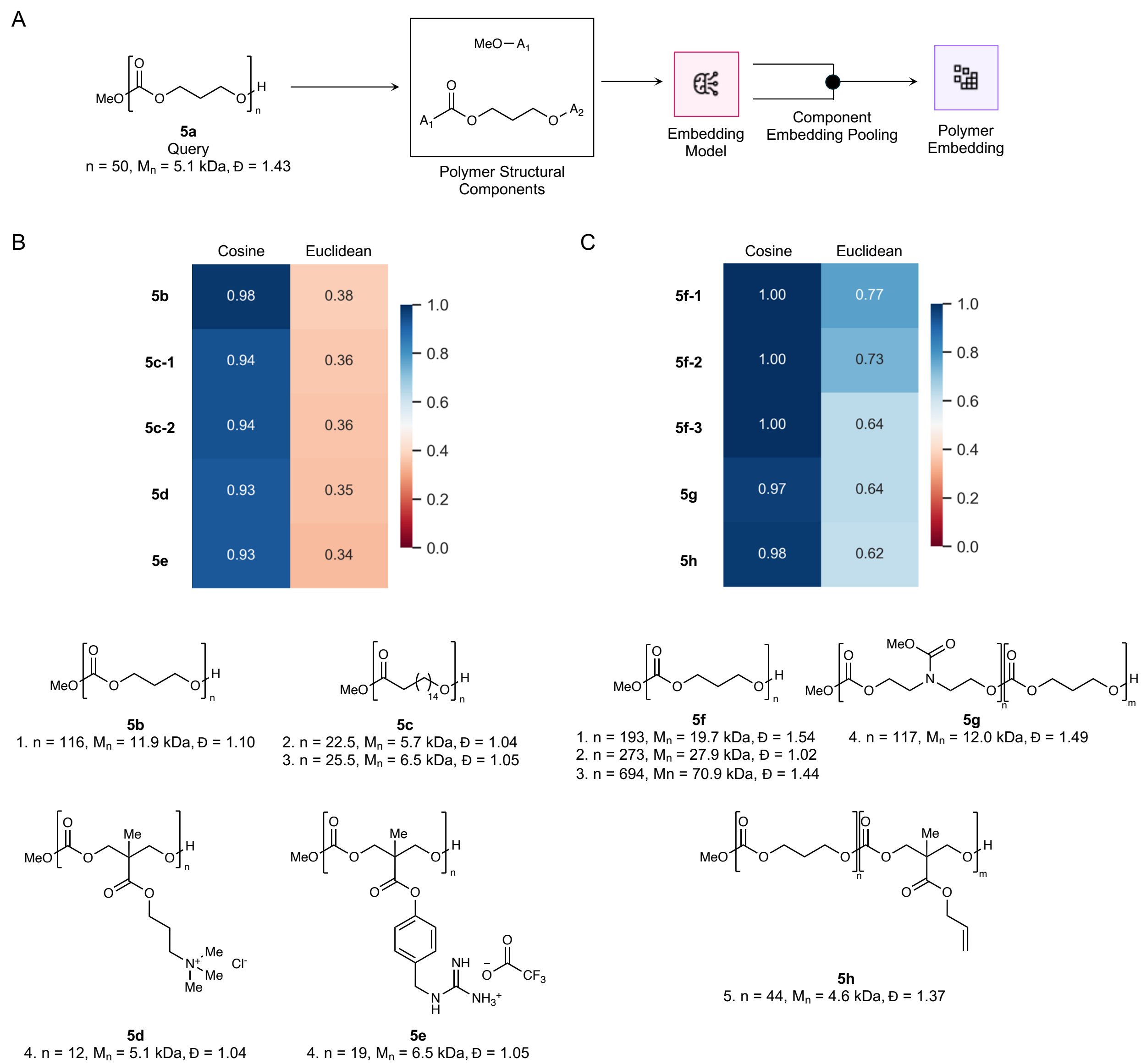

**Figure 3**. A) Schema of how an example query polymer **5a** is broken into structural components and then processed into single embedding to facilitate querying across both polymer structure and other characteristics such as $M_n$ B). Results from query of **5a** embedded using equation S2. C) Results from query of **5b** embedded using equation S3.

As with polymers, reactions—represented using reaction SMILES syntax[40]—has not been examined directly using MoLFormer and despite the existence of many transformer-based models for reactions,[33,41] we were interested in probing the versatility of MoLFormer embedding model reaction similarity queries. Based on our results with polymeric systems, we anticipated reactions to behave in a similar manner given a similar syntactical construction. The data for the evaluation was sourced from an open ~2M reaction dataset sourced from the USPTO used previously in transformer models for reactions, open publications, and historical experimental polymerization reaction data.[42] In addition to querying on whole reactions, we also focused on queries involving one to two reagents (Fig. S4A). In these examples, **S5a**, **S5b**, and their average vector where each used to search the database and provided reaction examples where structurally related reagents were used or the query molecules themselves (Fig. S4A). Finally, testing the order of **S5a** and **S5b**, either in a dot separated substrate series or on either side of angle brackets (>>) separating substrates and products produced markedly different results, indicating the importance of order within the reaction SMILES sequence. The difference in results when angle brackets are used is understandable considering that this would indicate two very different reactions when the order is reversed. However, unlike with polymers, the distinction between the ordering of reactants or products in reaction SMILES has less relevance unless some order of addition is being encoded. In total, these results indicate that the base model of MoLFormer is well-suited to capture structural relationships among small-molecules, polymers, and reactions.

The demonstration of MoLFormer as an effective embedding model beyond simple small molecules prompted us to examine the use of chemical structural embeddings with additional modalities. Association of chemical structure with different types of characterization data is a highly important task for any project within chemistry and materials domains. While characterization data is available in many formats, we opted to leverage the data pre-plotted and saved as images as input for creation of their corresponding vector embeddings. The ability to query available characterization data by either image or structure would be a powerful addition to both traditional data infrastructures and RAG pipelines. Implementing this task would typically require either some form contrastive learning or latent fusion, necessitating fine-tuning of an existing model or training of an additional model. Instead, we opted to test alternative strategies where the chemical components of a particular piece of characterization data would be embedded using MolFormer while the image components would be embedded using OpenCLIP (see Supplementary Information for details).[43,44] While CLIP models for images can also co-embed text captions, it is not anticipated these embeddings would be able to effectively capture nuances in chemical structure as a chemistry focused model like MoLFormer. Instead, the text embedder of the OpenCLIP model can be used to embed information captured as a natural language caption, which can be automatically generated from the metadata of characterization data files, adding additional natural language query capabilities. To evaluate this approach, we compiled a small dataset of labeled images of $^{1}$H, $^{13}$C, and $^{19}$F NMR spectra (see Supplementary Information). The labels for the spectra included their corresponding chemical components as SMILES strings and natural language captions generated from the spectra metadata. From this dataset we can test both image-based and structure-based queries on the embedded dataset with excellent results across numerous collection organization strategies (Fig. 4). Despite there being only minor differences between the image features plots—colors, whitespace, axes, and image size—of the embedded plots, OpenCLIP can effectively differentiate between different characterization techniques without additional metadata filtering on the query. This implies that the differentiation is likely occurring due to shape of the plotted traces or spectra (Fig. 4).

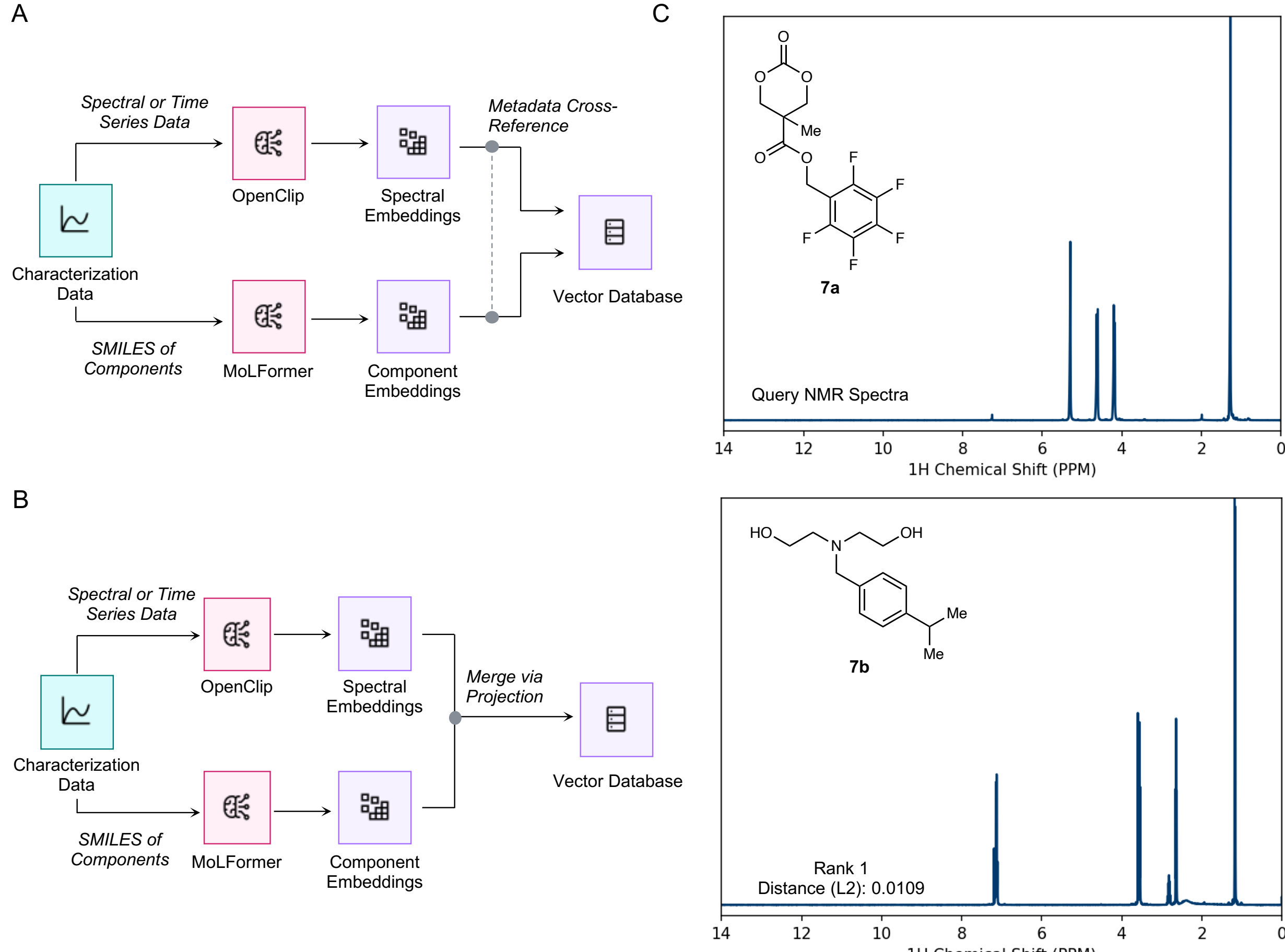

**Figure 4**. A) Schematic of embedding process for characterization data where the image and structural components share a metadata reference. B) Schematic of embedding process for characterization data where the image and structural embeddings are projected into a single vector before storage in a vector database. C) Example of query 1H NMR spectra and the top rank result from a querying against vector store built using cross-referenced metadata approach.

The use of chemistry LLMs such as MoLFormer and multi-modal models like OpenCLIP, in combination with post-embedding pooling and compression strategies, has facilitated the creation of a variety of vector stores which in turn can support a large variety of similarity queries. This alone would make a powerful addition to traditional database architectures and search capabilities, yet it's the connection of such vector databases within larger agentic RAG workflows can offer the potential of significant time savings in complex research tasks requiring the merging and summarization of data retrieved from complex structure, image, and natural language-based queries. In this regard, access to different vector stores and their respective embedding strategies is provided to LLM agents as tools which may be used in the context of a particular task. We utilized the LangGraph library (v 0.1.19)[45] to develop a hierarchical multi-agent adaptive self-reflective RAG system (Fig. 5A and 5B), which takes a question as input and outputs, the solution as a formatted report. As shown in Fig. 5, the hierarchical workflow is directed by a supervisor agent which leverages query analysis[46] to adaptively route a user's question to the correct worker agent. The system consists of four worker agents, which specialize in small molecules, polymers, chemical reactions, or NMR spectra. Each of these workers autonomously implement a multi-agent-based self-reflective RAG workflow.[47] As detailed in Fig. 5B, the workflow consists of several steps including retrieving and evaluating documents, generating responses using retrieved documents, revising user input to improve retrieved documents, checking generated responses for hallucinations and verifying each response completely addresses a user's question. Each step of this process is autonomously directed by the worker agents, which runs until all checks are passed. Once all checks are passed, the answer is sent to the report generation tool, which summarizes the agent's findings (Fig. S5 and Fig. S6). This report also summarizes the content retrieved from the vector store used by the agent to generate the report in order to improve the overall transparency of the workflow.

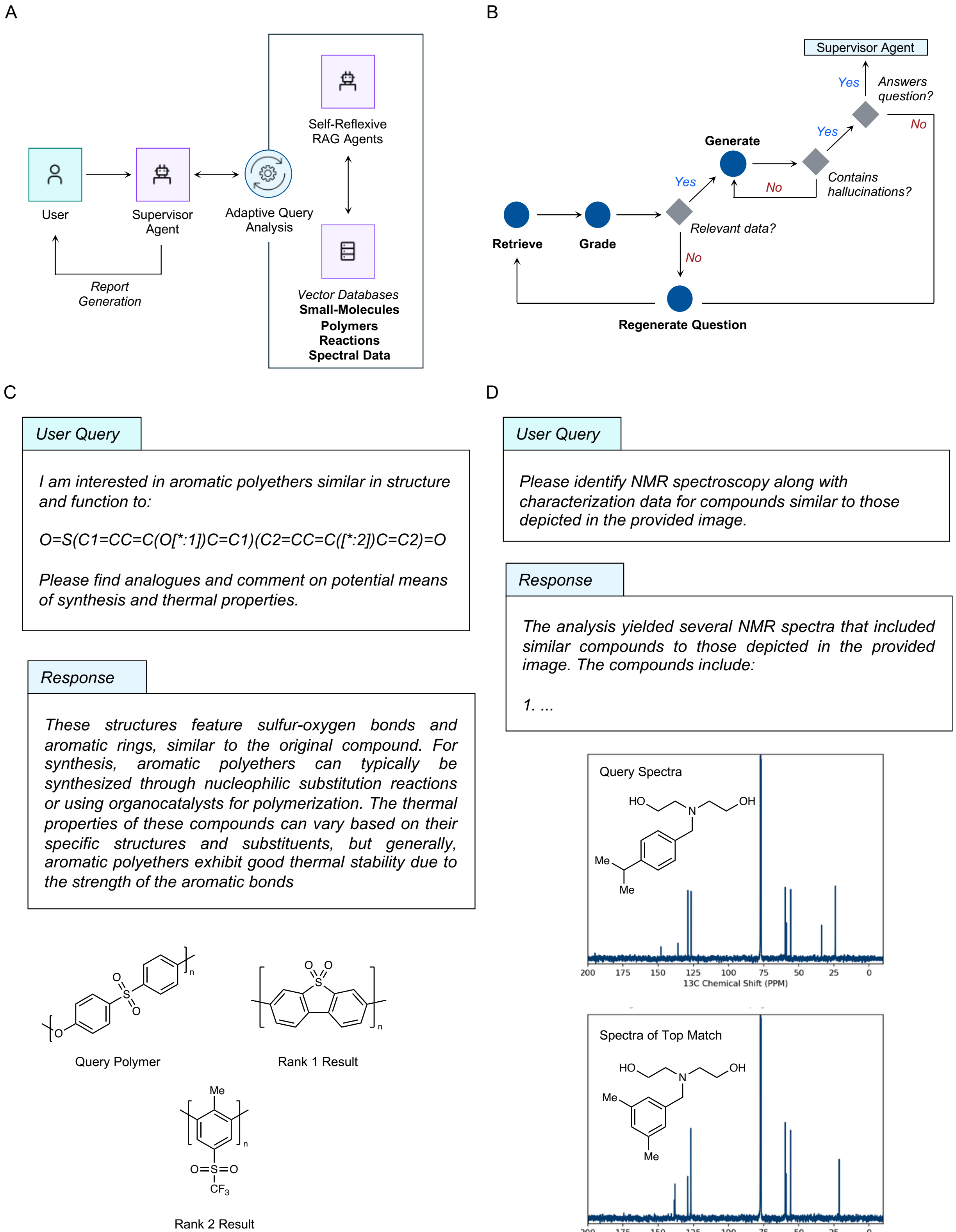

**Figure 5**. A) Schematic overview of hierarchical multi-agent system RAG to handle queries for different collections. B) Self-reflective RAG pipeline used in the adaptive query analysis by each RAG worker agent. C) Summarized

example of a polymer query and response from the multi-agent system. D) Summarized example of an NMR spectrum query and response from the multi-agent.

This workflow was demonstrated on a variety of tasks, each of which was able to effectively utilize the structure and/or image-based vector store tools in combination with other natural language-based vector databases from historical manuscripts. Summarized examples are shown in Fig. 5C and 5D with the full reports shown in Supplementary Figs. 5 and 6. As shown in Fig. 5C, the multi-agent workflow was able to successfully complete the user query using self-reflective RAG for polymers returning two similar analogs and provided additional content to justify the selection of these analogs. It should be noted that the collections utilized for the evaluation with the multi-agent system were limited to 250k examples from the original collection, thus covering a narrower range of compound types. Regardless, the structure-based RAG approach was able to retrieve close analogues for further utilization by the agents. For Fig. 5D, the agentic system was asked to identify similar $^{13}$C NMR spectra to the one of a diethanolamine-based carbonate monomer precursor.[49] The question was routed to the multi-modal NMR self-RAG agent who reviewed and summarized the characterization data of four $^{13}$C NMR spectra images (three retrieved images and the image corresponding to the input) summarizing the location of the peaks in these images that are similar to the input NMR spectra. The report also provides a visualization of the input image, which enables easier review and interpretation of the provided results. Both reports were reviewed by a domain expert who confirmed the validity of the findings. In this example, the agent was not instructed to filter the RAG results if they contained the original query image, and was still able to successfully find several spectra of closely related diethanolamine-based cyclic carbonates or their precursors (Fig. 5C, Fig. S6).

## Conclusion

The accelerated development of novel materials and catalysts necessitates dramatic improvement of human–AI interactions to facilitate both effective co-designs as well as realistic implementations within an experimental setting. Multi-agent RAG systems integrated with chat interfaces hold significant promise to become useful assistants for researchers operating in laboratory settings. Here, we have demonstrated that chemical foundation models coupled with powerful Multi-modal models can facilitate unique types of structural or architectural focused queries on small-molecules, polymers, and reactions. This represents a significant enhancement of search capabilities not typically found in traditional database systems for chemistry research. Moreover, the coupling of these systems within LLM-based multiagent system can provide a significant advantage for reducing the time needed to retrieve and summarize relevant multimodal structure-based information commonly required across all material research projects.

## Code and Data Availability

Data and code will be made available upon final publication.

## Conflict of Interest

A patent application on aspects of multimodal semantic search for materials has been filed by IBM (application number 18/798970, inventors: N.H.P, T.E., and J.L.H). All authors are employees of IBM.

## Acknowledgements

This material is based upon work supported by the NSF under Grant No. DBI-1548297.

# Supplementary Information

## Materials and Methods

*Vector Database Setup.* All SMILES embeddings were computed using the *ibm/MoLFormer-XL-both-10pct* model available through HuggingFace using canonicalized SMILES (with the exception of the USPTO dataset, which was used as is).[50] For the example queries in Figures 1-4, the vector embeddings for each compound were L2 normalized prior to insertion into a Milvus lite vector database using and Hierarchical navigable small world (HNSW) index and the L2 distance as the metric. For the multimodal examples (Fig. 5) SMILES embeddings were not L2 normalized and used in a Milvus standalone vector database using IVF_FLAT index and the L2 distance metric. For spectra containing more than one identifiable compound, including the NMR solvent, their corresponding embeddings were averaged prior to insertion. Image embeddings were computed using OpenCLIP via the Langchain library. The model used was *ViT-g-14* with the *laion2b_s34b_b88k* checkpoint. Image embeddings were stored in a separate Milvus collection, using an IVF_FLAT index and L2 distance metric, and cross-referenced with their corresponding compound embeddings.

*Similarity Metrics*. For small molecule and polymer example queries, similarity metrics were computed on the basis of both the MoLFormer embeddings and molecular fingerprints computed using RDKit. Cosine similarity was computed by subtracting the cosine distance between the query and result embeddings (measured using the SciPy library) from 1. Euclidean similarity ($E_s$) was computed by the following equation:

$$E_s = 1 / (1 + E_d) \tag{1}$$

Where $E_d$ is the Euclidean distance between the query and result embeddings. Tanimoto and Dice similarities were computed using the corresponding query and result Morgan fingerprints with a radius of 2 and a dimension of 2048. RDKit similarity was computed using the built-in RDKit fingerprints of the query and compound. Molecular ACCess System (MACCS) similarity was computed using the MACCS Keys fingerprints of the query and the compound using the RDKit package.

*Language Agent Network*. As noted in the main text, the multi-agent framework was assembled using LangChain and LangGraph libraries. The supervisor agent utilized *GPT-4o mini* model while other agents used either *llava 7b* or *llama3.1 8b* models. All Milvus collections were instantiated as separate vector stores with customized embedding functions with either MoLFormer or OpenCLIP, prior to connection with LLM agents as retrievers. Full code for agent network will be released in both subsequent drafts of preprint and final publication.

*Synthetic Polymer Data Generation.* A synthetic dataset of homopolymers and AB block copolymers of polycarbonates, polyesters, polysiloxanes, polyacrylates, and polystyrenes was created through enumerative combination of compatible monomers and initiators pulled from historical datasets. Values for $M_n$, $DP_n$, and dispersity where randomly generated within specific ranges and assigned to the enumerated polymer structure. The resulting dataset contained >1M synthetic polymers with attendant molecular weight properties.

## Supplementary Figures

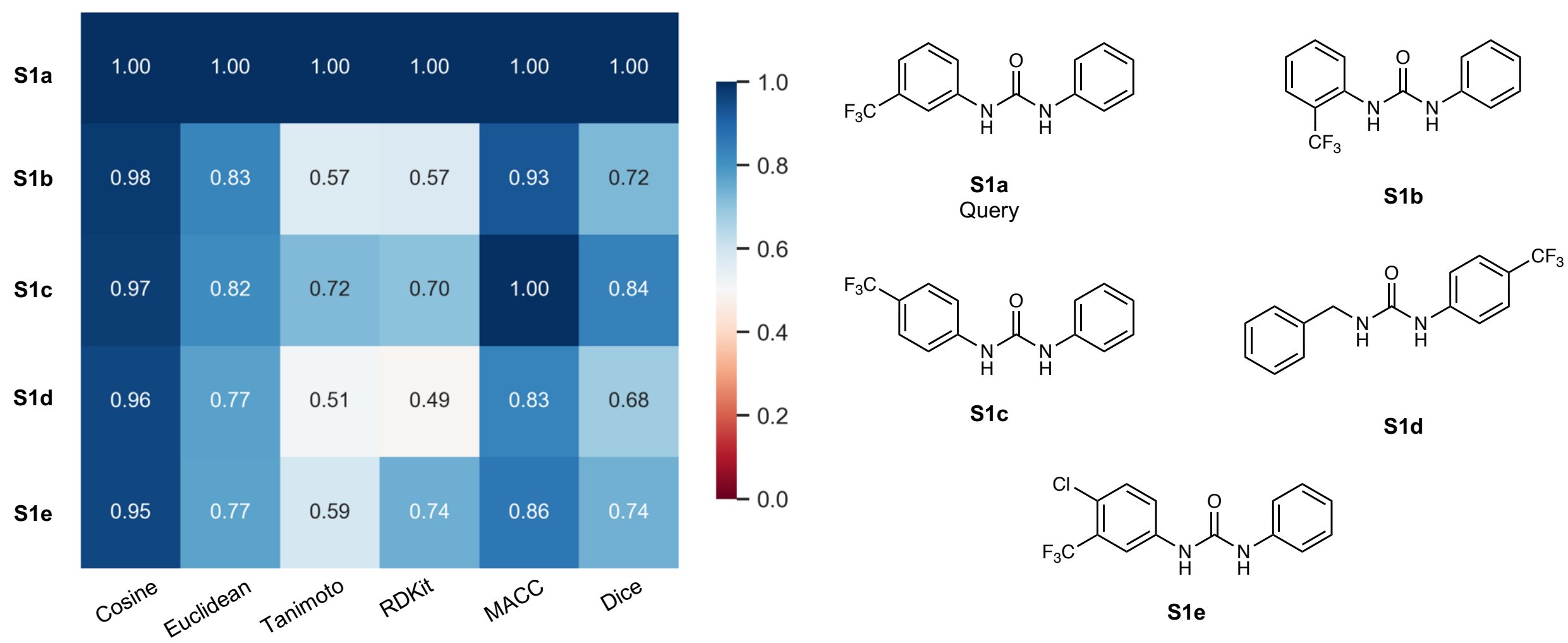

**Figure S1**. Additional example of small-molecule similarity query.

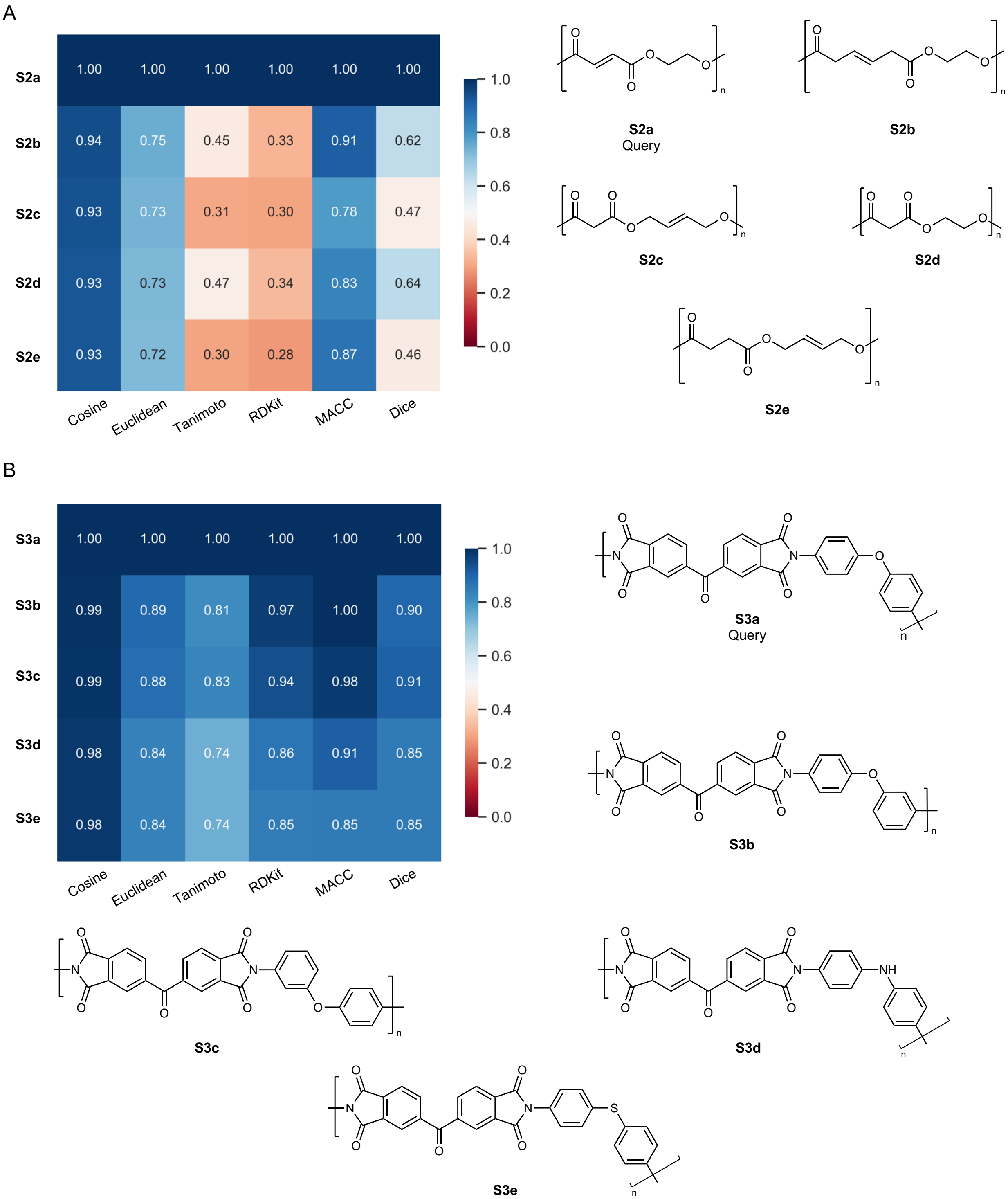

**Figure S2**. Examples of similarity searches for polymers based on PSMILES strings.

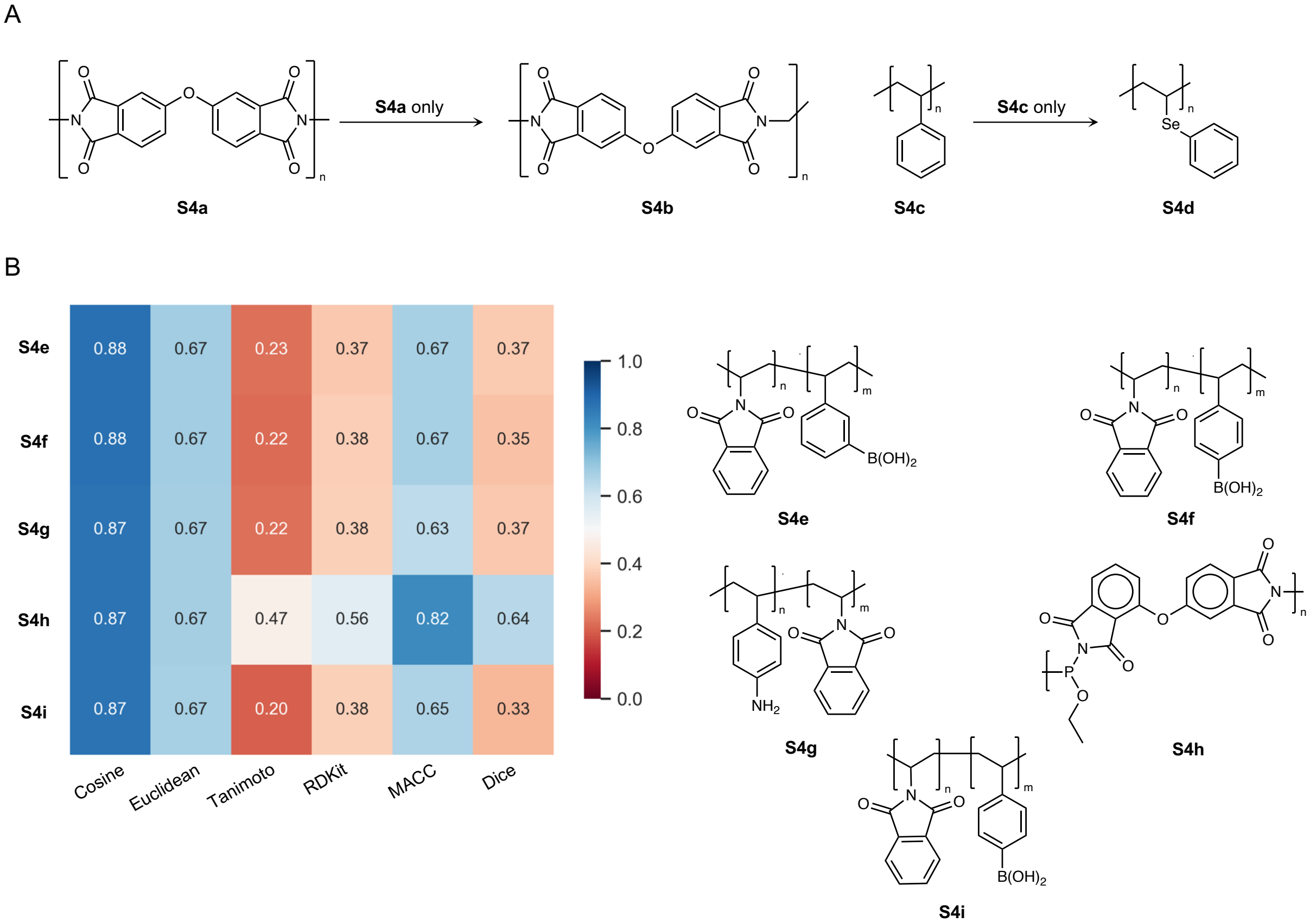

**Figure S3**. A) Examples of similarity searches for polymers based on PSMILES strings. B) Results of a query on the average embedding of **S4a** and **S4b**.

A

R = H

R =

R =NO2

R = NH2

Query: **S5a**

Query: **S5b**

Query: average of **S5a** and **S5b**

B

Query: **S5a.S5b** and **S5a>>S5b**

Query: **S5b.S5a**

Query: **S5b>>S5a**

**Figure S4.** A) Top results of queries against 2.5 M USPTO reaction collection of embedded reaction SMILES for either **S5a**, **S5b**, or their average. B) Comparison of top results from concatenated query SMILES strings of **S5a** and **S5b** either dot separated, such as in **S5a**. **S5b**, or angle-bracket separated, such as in **S5a**>> **S5b**. Other reagents and solvents are excluded for clarity.

```
INTRODUCTION
------------
The user is interested in finding analogues of a specific aromatic polyether and exploring potential synthesi
s methods and thermal properties.
The user's exact query was: 'I am interested in aromatic polyethers similar in structure and function to O=S
(C1=CC=C(O[*:1])C=C1)(C2=CC=C([*:2])C=C2)=O. Please find analogues and comment on potential means of synthesi
s and thermal properties.'.

Input Compound: O=S(C1=CC=C(O[*:1])C=C1)(C2=CC=C([*:2])C=C2)=O
```

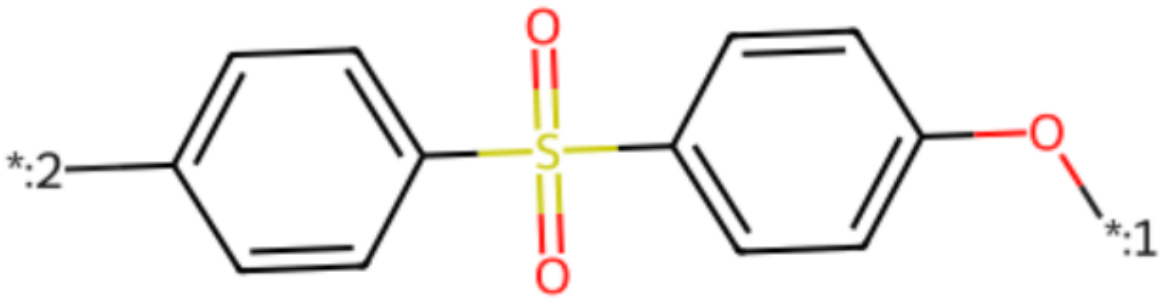

```
RESEARCH STEPS
--------------
I utilized the polymer_rag_search tool to identify aromatic polyethers similar in structure and function to t
he provided SMILES string. The tool returned two relevant SMILES strings that represent potential analogues.

MAIN FINDINGS
-------------
The identified analogues are:
1. [*:1]c1ccc2c(c1)S(=O)(=O)c1cc([*:2])ccc1-2
2. [*:1]c1cc(S(=O)(=O)C(F)(F)F)cc([*:2])c1C

These structures feature sulfur-oxygen bonds and aromatic rings, similar to the original compound. For synthe
sis, aromatic polyethers can typically be synthesized through nucleophilic substitution reactions or using or
ganocatalysts for polymerization. The thermal properties of these compounds can vary based on their specific
structures and substituents, but generally, aromatic polyethers exhibit good thermal stability due to the str
ength of the aromatic bonds.

CONCLUSION
----------
In summary, two potential analogues of the specified aromatic polyether have been identified, along with gene
ral methods for their synthesis and considerations regarding their thermal properties.

SOURCES
-------
- polymer_rag_search tool for identifying analogues and synthesis methods.

**Summary of Vector Store Context:**
Compound(s):
         - [*:1]c1cc(S(=O)(=O)C(F)(F)F)cc([*:2])c1C)
```

**Figure S5**. Example of a report generated for a question routed to the self-reflective RAG agent for polymers.

```
INTRODUCTION
------------
The user requested identification of NMR spectroscopy images and characterization data for compounds similar to those depicted in a provided image.
The user's exact query was: 'Please identify NMR spectroscopy images along with characterization data for compounds that are similar to those depicted in the provided image. data/images/NMR/83a-13C/83a-1.png'.

Input Image:
```

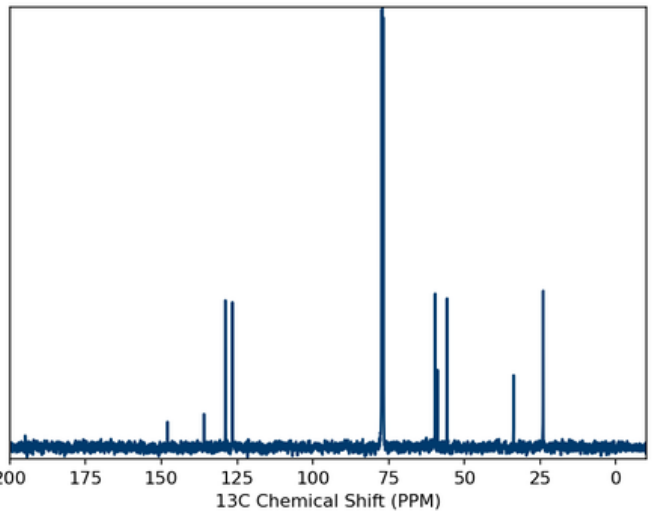

```
RESEARCH STEPS
--------------
I utilized the nmr_image_rag_search tool to analyze the provided image and retrieve relevant NMR spectra and characterization data for similar compounds.

MAIN FINDINGS
-------------
The analysis yielded several NMR spectra and their corresponding peaks for compounds similar to those depicted in the provided image. The compounds include:
1. **CC(C)c1ccc(CN(CCO)CCO)cc1** - Peaks (ppm): 148.0, 135.9, 128.9, 126.6, 77.4, 77.0, 76.7, 59.6, 58.9, 55.7, 33.8, 24.0.
2. **Cc1cc(C)cc(CN(CCO)CCO)c1** - Peaks (ppm): 138.6, 138.0, 129.0, 126.8, 77.4, 77.0, 76.7, 59.6, 59.1, 55.8, 21.3.
3. **Cc1cc(C)c(CN2CCOC(=O)OCC2)c(C)c1** - Peaks (ppm): 138.2, 137.1, 130.8, 129.2, 77.9, 77.8, 77.7, 77.6, 77.4, 77.0, 76.7, 69.8, 54.7, 53.8, 20.9, 20.0.
4. **OCCN(CCO)C1c2ccccc2-c2ccccc21** - Peaks (ppm): 144.4, 140.8, 128.4, 127.4, 125.2, 120.1, 78.1, 77.7, 77.4, 77.0, 76.7, 67.8, 60.5, 54.1.

CONCLUSION
----------
The retrieved NMR spectra and characterization data provide valuable insights into compounds similar to those depicted in the provided image, highlighting their structural features and chemical environments.

SOURCES
-------
- NMR spectra and characterization data retrieved from the nmr_image_rag_search tool.

**Summary of Vector Store Context:**
Compound(s):
        - Cc1cc(C)cc(CN(CCO)CCO)c1;ClC(Cl)Cl
        - Cc1cc(C)c(CN2CCOC(=O)OCC2)c(C)c1;ClC(Cl)Cl
        - OCCN(CCO)C1c2ccccc2-c2ccccc21;ClC(Cl)Cl

The NMR Spectra of the most relevant compound is shown below
(data/images/NMR/83b-13C/83b-1.png):
```

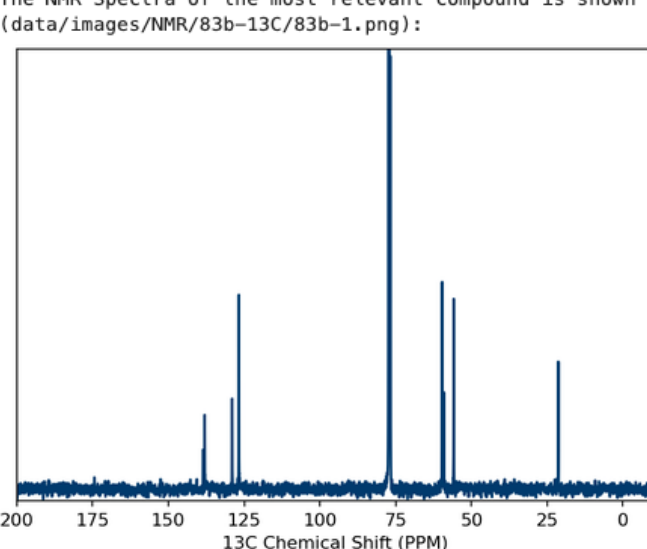

```
Figure Caption: A 100.62 MHz 13C NMR in CDCl3 collected on 2021-02-02T14:25:28 with 256
scans and a scan delay of 4 seconds. The displayed spectrum is between 0 and 12 PPM. Peaks
(ppm): 138.6, 138.0, 129.0, 126.8, 77.4, 77.0, 76.7, 59.6, 59.1, 55.8, 21.3
```

**Figure S6.** Example of report generated from an image query of a 13C NMR.